\documentclass{article}

\usepackage[main, final]{neurips_2026}
\makeatletter
\renewcommand{\@notice}{}
\makeatother

\usepackage[utf8]{inputenc} 
\usepackage[T1]{fontenc}    
\usepackage{hyperref}       
\usepackage{url}            
\usepackage{booktabs}       
\usepackage{amsmath}        
\usepackage{amsfonts}       
\usepackage{amssymb}        
\usepackage{nicefrac}       
\usepackage{microtype}      
\usepackage[table]{xcolor}  
\usepackage{graphicx}       
\usepackage{subcaption}     
\usepackage{multirow}       
\usepackage{tabularx}

\usepackage{enumitem}

\title{SimpleVLN: Towards Simple and Frontier Vision-Language Navigation}

\author{%
  \textbf{Lingyi Hong}$^{1,*,\dagger}$, \textbf{Weiren Cai}$^{2,*}$, \textbf{Jiayu Chen}$^{1,*}$,  \textbf{Xinyu Zhou}$^{2,\dagger,\ddagger}$, \\
  \textbf{Bailin He}$^{1}$, \textbf{Yunfei Guo}$^{1}$, \textbf{Qiao Sun}$^{3}$, \textbf{Boyu Ma}$^{2}$, \\ 
  \textbf{Jianfei Yang}$^{2}$, \textbf{Wenqiang Zhang}$^{1}$ \\[4pt]
  $^{1}$\textnormal{Fudan University}, $^{2}$\textnormal{Nanyang Technological University}, $^{3}$\textnormal{Lenovo Research} \\[4pt]
  \texttt{\href{https://simplevln.github.io/}{\textcolor{magenta}{Project Homepage}}} \\
  \texttt{honglyhly@gmail.com}, \texttt{xinuzprf@gmail.com}
}

\begin{document}

\maketitle

\begingroup
\renewcommand{\thefootnote}{\fnsymbol{footnote}}
\footnotetext[1]{Equal contribution.\quad $^{\dagger}$Joint project leaders.\quad $^{\ddagger}$Corresponding author.}
\endgroup

\begin{abstract}

Vision-and-Language Navigation (VLN) requires embodied agents to ground language in visual observations and make long-horizon decisions. Existing methods often address these demands through task-specific architectures. As multimodal large language models (MLLMs) become increasingly capable, an important question remains: \textbf{\textit{Have we truly unlocked the potential of native MLLMs for VLN before introducing task-specific architecture?}} We present \textbf{SimpleVLN}, an architecture-preserving framework developed through studies of navigation formulation, data construction, and optimization. At the formulation level, SimpleVLN combines single-turn decision inputs with fixed-budget global--local memory to efficiently use navigation history without unbounded context growth, while action chunks enable efficient output of multiple primitive actions per model invocation. For data construction, we combine expert demonstrations with policy-dependent DAgger experience and develop a scalable synthesis pipeline that generates large-scale tasks for reinforcement learning. At the optimization level, a progressive IL--DAgger--RL pipeline advances from action imitation to task-level outcome optimization. Comprehensive experiments validate the effectiveness of these three aspects. We instantiate SimpleVLN across multiple MLLM backbones and model sizes, demonstrating its generality. SimpleVLN achieves $76.7\%$ and $80.1\%$ success rates on the R2R-CE and RxR-CE splits, respectively, substantially outperforming existing methods. These results show that the navigation capabilities of native MLLMs can be effectively unlocked without task-specific architectural designs.

\end{abstract}

\section{Introduction}

Vision-and-Language Navigation (VLN) requires embodied agents to follow natural-language instructions in unseen environments~\citep{r2r,rxr}, combining visual understanding, language grounding, spatial reasoning, and long-horizon decision making. Existing methods have advanced through task-specific designs for perception and spatial reasoning~\citep{janusvln}, graph-based modeling~\citep{chen2022think}, and hierarchical or dual-system decision making~\citep{dualvln}. These architectures are effective, while multimodal large language models (MLLMs) offer strong general-purpose visual understanding, reasoning, and instruction following~\citep{qwen3vl}. However, the performance of augmented MLLM systems reflects both the backbone and task-specific components, leaving the capability of a native MLLM alone unclear. This raises an underexplored question: \textit{Have we truly unlocked the potential of native MLLMs for VLN before introducing task-specific architecture?} To answer it, we investigate how far native MLLMs can be advanced for VLN without architectural modifications.

\begin{figure}[!t]
\centering
\includegraphics[width=0.98\textwidth]{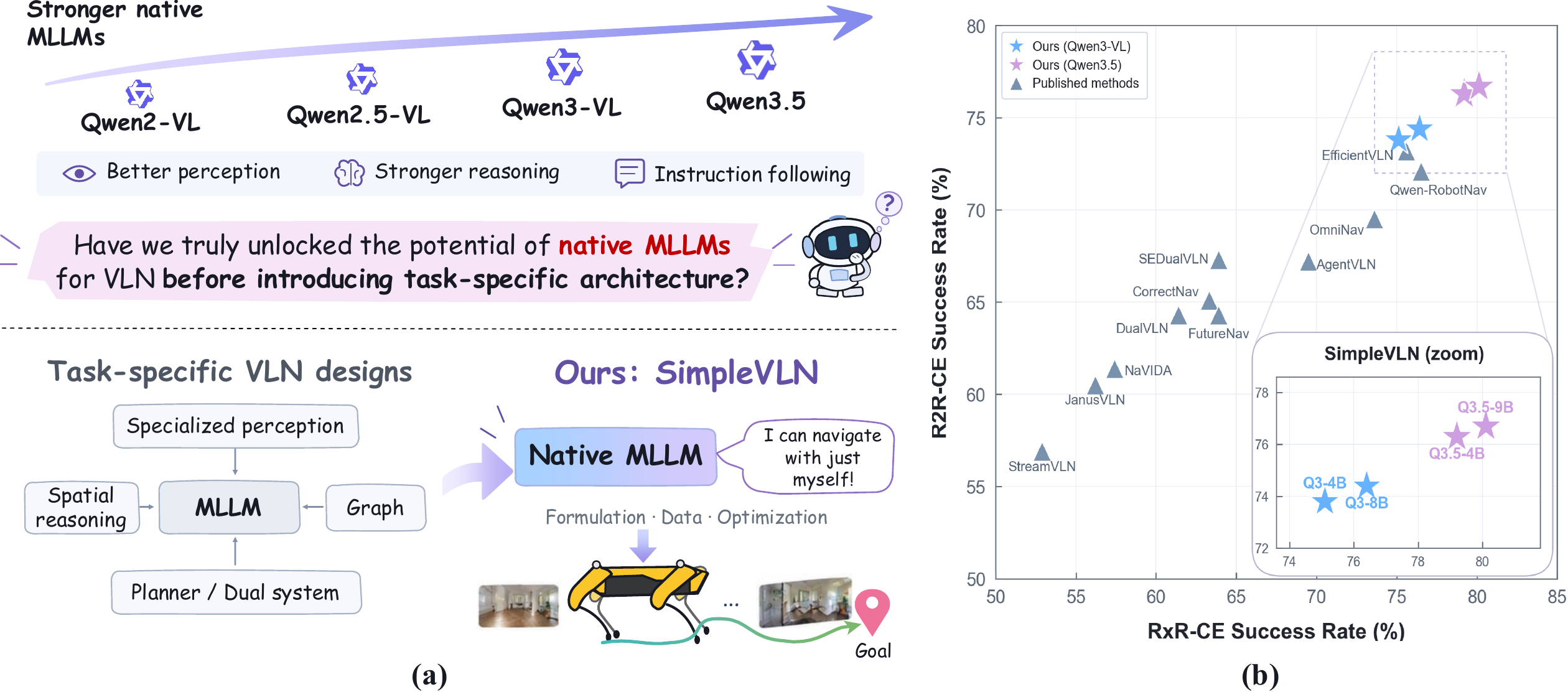}
\vspace{-2mm}
\caption{\textbf{Motivation and performance comparison of SimpleVLN}. (a) Motivation for using native MLLMs in VLN. (b) Comparison with existing methods on R2R-CE and RxR-CE.}
\label{fig:teaser}
\vspace{-5mm}
\end{figure}

Preserving the architecture still requires adapting the MLLM to sequential navigation. We develop \textbf{SimpleVLN} by studying three aspects: formulation, data construction, and optimization (Figure~\ref{fig:teaser}). First, at the formulation level, we combine single-turn decision inputs with fixed-budget global--local memory to prevent unbounded context growth while efficiently using navigation history. Action chunking further enables efficient action output by predicting multiple primitive actions per model invocation. Second, for data construction, we combine expert demonstrations with policy-dependent experience collected through DAgger~\citep{dagger}. We further develop a scalable task-synthesis pipeline to generate large-scale navigation tasks for reinforcement learning. Third, at the optimization level, we investigate a progressive IL--DAgger--RL training pipeline. IL initializes the policy through action imitation, DAgger adapts it to policy-induced states, and RL directly optimizes task-level outcomes, progressively improving navigation accuracy.

Comprehensive experiments validate the effectiveness of these three aspects. We instantiate SimpleVLN across multiple MLLM backbones and model sizes, demonstrating its generality. SimpleVLN achieves $76.7\%$ and $80.1\%$ SR on the R2R-CE and RxR-CE \textit{val-unseen} splits, respectively, substantially outperforming existing methods. These results demonstrate that the navigation capabilities of native MLLMs can be effectively unlocked without task-specific architectural designs.

Our contributions are threefold:
(1) We introduce SimpleVLN, an architecture-preserving framework that adapts native MLLMs to VLN through single-turn decision inputs, managed visual history, and action-chunk prediction, without task-specific architectural modules.
(2) We construct training data from expert demonstrations, policy-dependent DAgger experience, and scalable RL task synthesis, and develop a progressive IL--DAgger--RL pipeline for improving navigation accuracy.
(3) We instantiate SimpleVLN across multiple backbone families and model sizes, achieving strong performance on R2R-CE and RxR-CE, while extensive experiments validate the effectiveness of its formulation, data construction, and optimization.

\section{Related Work}

\noindent\textbf{Vision-and-Language Navigation.}
Vision-and-Language Navigation (VLN) requires agents to follow natural-language instructions grounded in visual observations~\citep{r2r,rxr}. While early VLN operates over predefined discrete viewpoints, VLN-CE~\citep{vlnce} extends navigation to continuous environments with low-level actions.

\noindent\textbf{Multimodal Large Language Models for VLN.}
Recent VLN methods leverage multimodal large language models (MLLMs) for vision--language understanding~\citep{navid,navila,uninavid}, while introducing streaming contexts~\citep{streamvln}, 3D geometry encoders~\citep{janusvln}, dual-system architectures~\citep{dualvln,sedualvln,agentvln,internvlan1}, or explicit chain-of-thought reasoning~\citep{navcot,Aux-Think} for long-horizon navigation. In contrast, SimpleVLN directly adapts a pretrained MLLM without architectural modifications, retaining compatibility with existing MLLM inference frameworks.

\begin{figure}[!h]
\centering
\includegraphics[width=0.98\textwidth]{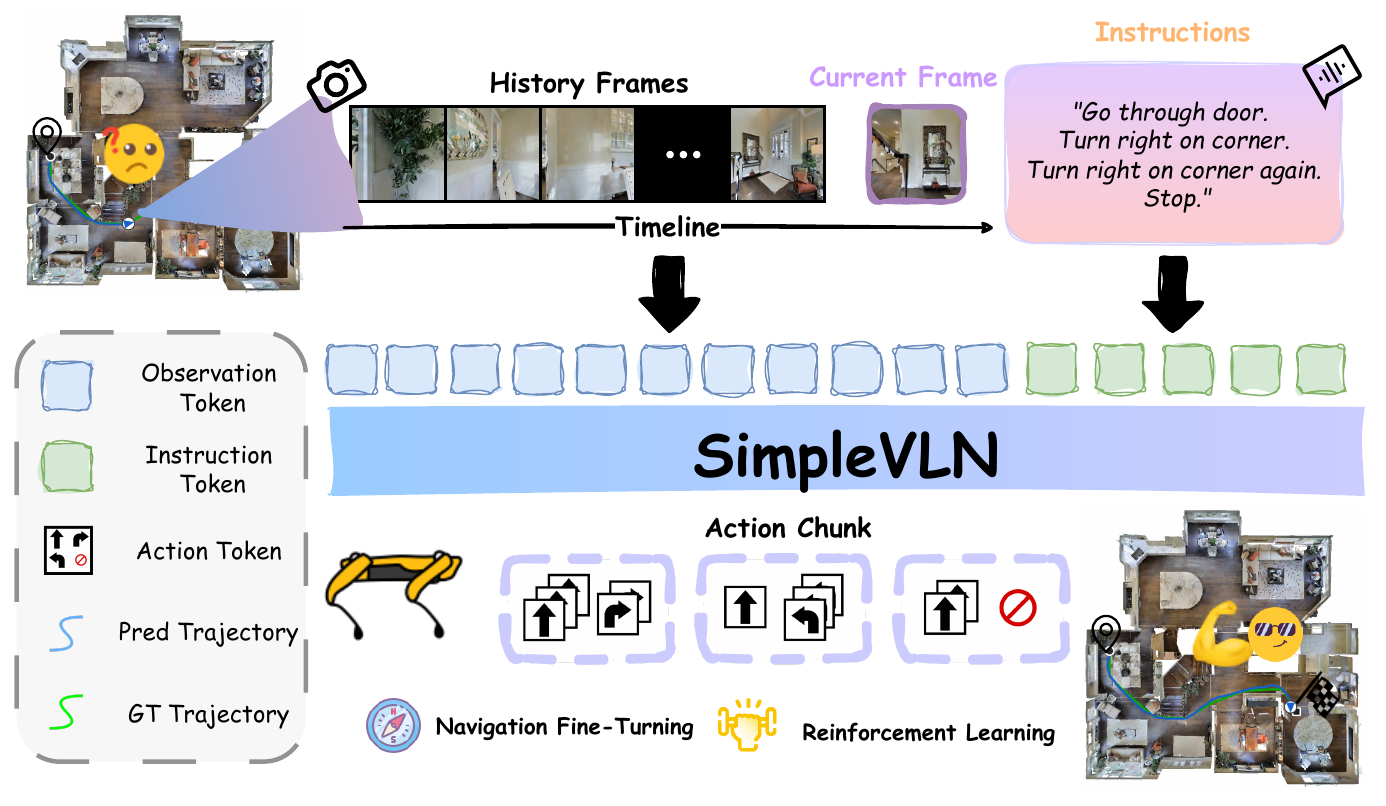}
\vspace{-2mm}
\caption{\textbf{Overview of SimpleVLN.} At each decision, sampled history frames, current observation, and the instruction are fed to an MLLM, which directly predicts an action chunk for navigation.}
\vspace{-5mm}
\label{fig:model}
\end{figure}

\noindent\textbf{Training Paradigms for Navigation.}
Imitation learning trains navigation policies from expert trajectories~\citep{streamvln,activevln}, but suffers from distribution shift under policy execution. DAgger~\citep{dagger} mitigates this issue using supervision on model-induced states~\citep{streamvln,janusvln}, while RL ~\citep{activevln,hu2026longnav} directly optimizes task-level outcomes. SimpleVLN combines them in a progressive IL-DAgger-RL pipeline from expert supervision to task-completion optimization.

\section{SimpleVLN}

\subsection{Overview}
\label{sec:overview}

SimpleVLN directly adapts off-the-shelf MLLMs~\citep{qwen35,qwen3vl} to VLN without architectural modifications. We study native-MLLM navigation from three perspectives: navigation formulation, data construction, and a progressive optimization strategy that shifts from action imitation to outcome-driven learning through environment interaction.

As illustrated in Figure~\ref{fig:model}, given a natural-language instruction $\mathcal{I}$, at each primitive environment step $k$, the agent observes an egocentric RGB image $x_k$ and executes an action $u_k$ from
\[
    \mathcal{A}
    =
    \{
    \textsc{Forward}(0.25\,\text{m}),
    \textsc{TurnLeft}(15^\circ),
    \textsc{TurnRight}(15^\circ),
    \textsc{Stop}
    \}.
\]
The episode terminates when $\textsc{Stop}$ is executed or the primitive-action budget $S_{\max}$ is reached, and succeeds if the agent stops within $3\,\text{m}$ of the target~\citep{vlnce}.

\subsection{Decision-Level Navigation Formulation}
\label{sec:formulation}

We formulate navigation around model decision steps rather than primitive environment steps. At each decision, SimpleVLN reconstructs a bounded context from the instruction, and sampled global--local history, and uses the native MLLM to predict navigation actions. We first describe the construction of the decision context and then introduce the output action representation.

\begin{figure}[t]
\centering
\includegraphics[width=\textwidth]{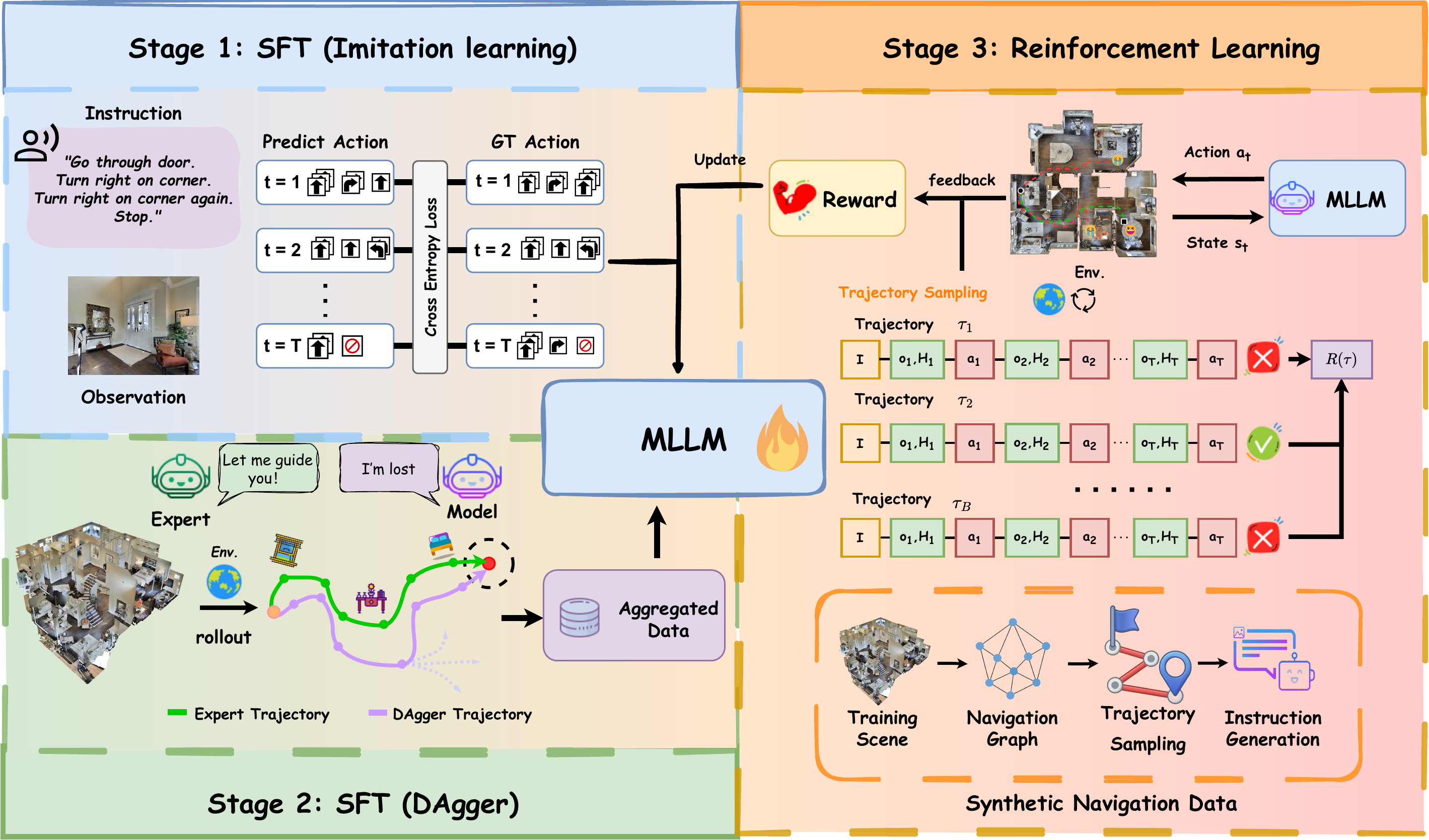}
\vspace{-2mm}
\caption{\textbf{Training data pipeline of SimpleVLN.} SimpleVLN first learns from expert demonstrations, then collects and aggregates corrective trajectories through DAgger. For RL, it synthesizes navigation episodes from sampled start--goal paths and generated instructions, then optimizes the policy through environment interaction and outcome feedback.}
\label{fig:pipeline}
\vspace{-5mm}
\end{figure}

\noindent\textbf{Single-turn decision input.}
At each model decision, we construct an independent input rather than append the latest observation and predicted action to the preceding dialogue.
Let $t=1,2,\ldots$ index model decisions, and let $n_t$ denote the number of primitive actions executed before decision $t$, with $n_1=0$. The input consists of the instruction $\mathcal{I}$, and a sampled visual history $H_t$; previous dialogue turns and past actions are not included.
We denote the resulting visual context by $s_t=H_t$.

\noindent\textbf{Global--local history selection.}
Before decision $t$, the primitive-step observation history $(x_0,\dots,x_{n_t})$ grows with environment interaction. Keeping all observations enlarges the input, whereas keeping only recent frames loses earlier route context. We therefore represent history at two complementary temporal scales: local history preserves recent continuity, while global history provides sparse coverage of earlier primitive-step observations. Let $L_t$ contain up to $N$ most recent observations, and let $G_t$ contain up to $M$ observations uniformly sampled from the earlier history:
\begin{equation}
\begin{aligned}
L_t = (x_k)&_{k=\max(0,n_t-N+1)}^{n_t}, 
\qquad m_t = \min\{M,\max(0,n_t-N+1)\}, \\
&G_t = (x_{i_j})_{j=1}^{m_t},
\qquad
H_t = G_t \mathbin{\Vert} L_t.
\end{aligned}
\label{eq:memory}
\end{equation}
Here, $\Vert$ denotes sequence concatenation. For $m_t>0$, the sampled indices satisfy
$0\le i_1<\dots<i_{m_t}<\max(0,n_t-N+1)$, preserving chronological order. When fewer than $M$ earlier observations are available, all are retained. Thus, each decision uses at most $N+M$ visual inputs, independently of episode length.

\noindent\textbf{Action chunking.}
On the output side, a single model decision can specify multiple primitive actions. Rather than predicting one primitive action per decision, we follow NaVIDA~\citep{navida} and predict an action chunk $a_t$, defined as an ordered sequence of primitive actions. Let $K_t$ denote the number of primitive actions in the chunk; then
$
a_t
=
(u_{n_t},\dots,u_{n_t+K_t-1}),$ and the navigation policy predicts
$
a_t
\sim
\pi_\theta
\big(
\cdot
\mid
\mathcal{I},s_t
\big)
,
\label{eq:chunk}
$
where $\pi_\theta$ denotes the native MLLM navigation policy. The environment executes the actions in $a_t$ sequentially, producing an observation after each. These observations are retained for later history construction. The MLLM is queried again when the chunk finishes, unless $\textsc{Stop}$ or the primitive-action budget terminates the episode.

\subsection{Navigation Data Construction}
\label{sec:synthetic_data}

We construct training data for each stage of post-training. Expert demonstrations provide action supervision, while DAgger adds policy-dependent states encountered during model execution. At the RL stage, reusing the same episodes used for supervised learning would not expand the task distribution or provide genuinely new interaction experience. We therefore synthesize new navigation tasks in the training scenes.

\noindent\textbf{Expert demonstration coverage.}
We pool trajectories from R2R~\citep{r2r}, RxR~\citep{rxr}, and ScaleVLN~\citep{scalevln} to broaden expert behavior coverage. To match the decision-level inference interface in Section~\ref{sec:formulation}, each trajectory is converted into supervised samples with $(\mathcal{I},s_t)$ as input and the corresponding expert action chunk $a_t^\star$ as the target, where $s_t=H_t$.

\noindent\textbf{Policy-dependent experience via DAgger.}
Broader expert demonstrations do not eliminate the distribution gap between expert trajectories and states visited by the current policy. We therefore use a deviation-based DAgger~\citep{dagger} strategy. We compare the current policy trajectory with the ground-truth (GT) trajectory: when the deviation is large, we select the expert trajectory; otherwise, we retain the policy trajectory. Selected trajectories are converted into decision-level supervised samples and aggregated with the demonstrations.

\noindent\textbf{Task synthesis for RL.}
Because the R2R, RxR, and ScaleVLN training episodes are already used for supervised learning, we synthesize new RL tasks in their training scenes rather than reuse the annotated episodes, excluding val-unseen scenes. For each source, we sample start--goal pairs in different rooms from the navigation graph, with goals associated with semantically labelled objects. An MLLM generates navigation instructions from reference-path waypoints describing turns, rooms, salient objects, and egocentric views. The reference path supports task construction rather than providing an action target.

\subsection{Progressive Optimization}
\label{sec:training}

Figure~\ref{fig:pipeline} summarizes the three-stage optimization of SimpleVLN. First, imitation learning uses expert demonstrations to learn navigation actions. Next, DAgger-based fine-tuning adapts the policy to states encountered during its own execution while retaining action-level supervision. Finally, reinforcement learning optimizes task-level outcomes through environment interaction on synthesized episodes, shifting the learning signal from demonstrated actions to outcome feedback.

\noindent\textbf{Imitation learning from expert actions.}
Let $\mathcal{D}$ denote a supervised dataset of tuples
$(\mathcal{I},s_t,a_t^\star)$, where $a_t^\star$ denotes the target action chunk for decision state $s_t$. We minimize the negative log-likelihood of the complete target chunk:
$
\mathcal{L}_{\mathrm{SFT}}
=
-\mathbb{E}_{(\mathcal{I},s_t,a_t^\star)\sim\mathcal{D}}
\left[
\log
\pi_\theta
\big(
a_t^\star
\mid
\mathcal{I},s_t
\big)
\right].
\label{eq:sft}
$
This objective teaches the policy to reproduce demonstrated action chunks, learning from expert supervision.

\noindent\textbf{Policy-dependent supervised adaptation.}
We next train on the aggregation of demonstration and DAgger data from Section~\ref{sec:synthetic_data}. The objective $\mathcal{L}_{\mathrm{SFT}}$ is unchanged, while $\mathcal{D}$ additionally includes target action chunks $a_t^\star$ derived from the selected policy-dependent trajectories.

\begin{table*}[t]
\caption{\textbf{Comparison with state-of-the-art methods} on the R2R-CE and RxR-CE \textit{val-unseen} splits. $^{\dagger}$ and $^{\ast}$ denote SimpleVLN before and after RL, respectively.}
\vspace{-1mm}
\label{tab:main}
\centering
\footnotesize
\setlength{\tabcolsep}{3.0pt}
\resizebox{\textwidth}{!}{%
\begin{tabular}{lc cccc cccc cccc}
\toprule
\multirow{2}{*}{Method} & \multirow{2}{*}{Size} &
\multicolumn{4}{c}{Observation} &
\multicolumn{4}{c}{R2R-CE val-unseen} & \multicolumn{4}{c}{RxR-CE val-unseen} \\
\cmidrule(lr){3-6}\cmidrule(lr){7-10}\cmidrule(lr){11-14}
 & & Pano. & Odo. & Depth & S.RGB
     & NE$\downarrow$ & OS$\uparrow$ & SR$\uparrow$ & SPL$\uparrow$
     & NE$\downarrow$ & SR$\uparrow$ & SPL$\uparrow$ & nDTW$\uparrow$ \\
\midrule
NaVid~\citep{navid}    & 7B   &            &            &            & \checkmark & 5.47 & 49.1 & 37.4 & 35.9 & 8.41  & 23.8 & 21.2 & ---  \\
Uni-NaVid~\citep{uninavid} & 7B &           &            &            & \checkmark & 5.58 & 53.3 & 47.0 & 42.7 & 6.24  & 48.7 & 40.9 & ---  \\
NaVILA~\citep{navila}  & 7B   &            &            &            & \checkmark & 5.22 & 62.5 & 54.0 & 49.0 & 6.77  & 49.3 & 44.0 & 58.8 \\
ActiveVLN~\citep{activevln} & 3B &        &            &            & \checkmark & 5.31 & 58.3 & 50.1 & 43.7 & 5.84  & 50.7 & 41.2 & 58.1 \\
StreamVLN~\citep{streamvln} & 7B &         &            &            & \checkmark & 4.98 & 64.2 & 56.9 & 51.9 & 6.22  & 52.9 & 46.0 & 61.9 \\
NavFoM~\citep{navfom} & 7B   &            &            &            & \checkmark & 5.01 & 64.9 & 56.2 & 51.2 & 5.51  & 57.4 & 49.4 & 60.2 \\
JanusVLN~\citep{janusvln} & 7B &           &            &            & \checkmark & 4.78 & 65.2 & 60.5 & 56.8 & 6.06  & 56.2 & 47.5 & 62.1 \\
NaVIDA~\citep{navida} & 3B &             &            &            & \checkmark & 4.32 & 69.5 & 61.4 & 54.7 & 5.23  & 57.4 & 49.6 & 67.0 \\
CorrectNav~\citep{correctnav} & 7B &      &            &            & \checkmark & 4.24 & 67.5 & 65.1 & 62.3 & 4.09  & 69.3 & 63.3  & 75.2 \\
DualVLN~\citep{dualvln} & 7B  &            &            &            & \checkmark & 4.05 & 70.7 & 64.3 & 58.5 & 4.58  & 61.4 & 51.8 & 70.0 \\
FutureNav~\citep{futurenav} & 4B &        &            &            & \checkmark & 4.24 & 70.2 & 65.4 & 61.3 & 4.60  & 60.9 & 52.8 & 68.0 \\
FutureNav~\citep{futurenav} & 8B &        &            &            & \checkmark & 4.29 & 69.2 & 64.3 & 60.3 & 4.26  & 63.9 & 54.8 & 69.1 \\
AgentVLN~\citep{agentvln} & 3B &          &            & \checkmark & \checkmark & 3.88 & 73.5 & 67.2 & 64.7 & 3.92  & 69.5 & 61.3 & 74.6 \\
OmniNav~\citep{omninav} & 3B & \checkmark &          &            &            & 3.74 & 74.6 & 69.5 & 66.1 & 3.77  & 73.6 & 62.0 & ---  \\
SEDualVLN~\citep{sedualvln} & 7B &         &            &            & \checkmark & 3.75 & 73.7 & 67.3 & 62.5 & 4.12  & 63.9 & 52.4 & 72.8 \\
Qwen-RobotNav~\citep{qwenrobotnav} & 4B & &            &            & \checkmark & 4.22 & 73.6 & 66.9 & 60.5 & 4.15  & 71.3 & 61.5 & 68.6 \\
Qwen-RobotNav~\citep{qwenrobotnav} & 8B & &            &            & \checkmark & 4.36 & 72.7 & 65.7 & 59.6 & 4.16  & 73.4 & 63.5 & 69.9 \\
Qwen-RobotNav~\citep{qwenrobotnav} & 4B & \checkmark & &          &            & 3.80 & 77.2 & 69.5 & 63.6 & 3.80  & 75.2 & 65.0 & 71.9 \\
Qwen-RobotNav~\citep{qwenrobotnav} & 8B & \checkmark & &          &            & 3.53 & 78.5 & 72.1 & 66.6 & 3.58  & 76.5 & 65.7 & 72.5 \\
Efficient-VLN~\citep{zheng2025efficient} & 8B &  & &          &  \checkmark  & 3.29 & 78.2 & 73.2 & 65.0 & 3.18 & 75.6 & 60.0 & - \\
\midrule
\rowcolor{gray!15}
\textbf{SimpleVLN$^{\dagger}$ (Qwen2.5-VL)} & 3B &            &            &            & \checkmark & 5.70 & 63.3 & 51.6 & 44.9 & 5.20 & 50.3 & 42.0 & 57.6 \\
\rowcolor{gray!15}
\textbf{SimpleVLN$^{\dagger}$ (Qwen2.5-VL)} & 7B &            &            &            & \checkmark & 4.87 & 68.4 & 59.1 & 52.0 & 5.31 & 58.9 & 49.2 & 62.9 \\
\rowcolor{gray!15}
\textbf{SimpleVLN$^{\dagger}$ (Qwen3-VL)} & 4B &            &            &            & \checkmark & 4.26 & 74.3 & 65.4 & 58.8 & 4.79 & 63.4 & 54.1 & 66.5 \\
\rowcolor{gray!15}
\textbf{SimpleVLN$^{\ast}$ (Qwen3-VL)} & 4B &            &            &            & \checkmark & 3.39 & 79.4 & 73.8 & 66.7 & 3.29 & 75.1 & 65.3 & 71.5 \\
\rowcolor{gray!15}
\textbf{SimpleVLN$^{\dagger}$ (Qwen3-VL)} & 8B &            &            &            & \checkmark & 4.40 & 73.3 & 64.2 & 57.0 & 4.73 & 64.2 & 54.3 & 66.2 \\
\rowcolor{gray!15}
\textbf{SimpleVLN$^{\ast}$ (Qwen3-VL)} & 8B &            &            &            & \checkmark & 2.93 & 79.2 & 74.4 & 69.5 & 3.11 & 76.4 & 66.4 & 77.4 \\
\rowcolor{gray!15}
\textbf{SimpleVLN$^{\dagger}$ (Qwen3.5)} & 4B &            &            &            & \checkmark & 3.76 & 77.5 & 67.1 & 59.2 & 4.32 & 65.6 & 55.3 & 67.8 \\
\rowcolor{gray!15}
\textbf{SimpleVLN$^{\ast}$ (Qwen3.5)} & 4B &            &            &            & \checkmark & 2.79 & 80.1 & 76.3 & 72.3 & 2.61 & 79.2 & 69.7 & 80.3 \\
\rowcolor{gray!15}
\textbf{SimpleVLN$^{\dagger}$ (Qwen3.5)} & 9B &            &            &            & \checkmark & 3.59 & 76.9 & 69.2 & 61.2 & 4.48 & 64.0 & 54.2 & 67.0 \\
\rowcolor{gray!15}
\textbf{SimpleVLN$^{\ast}$ (Qwen3.5)} & \textbf{9B} &            &            &            & \textbf{\checkmark} & \textbf{2.63} & \textbf{84.1} & \textbf{76.7} & \textbf{72.6} & \textbf{2.51} & \textbf{80.1} & \textbf{69.9} & \textbf{79.8} \\
\bottomrule
\end{tabular}}
\vspace{-3mm}
\end{table*}

\noindent\textbf{Outcome optimization through reinforcement learning.}
Because navigation is ultimately evaluated by task completion rather than agreement with a particular action sequence, we optimize task-level outcomes on the synthesized navigation tasks. Using the synthesized episodes described in Section~\ref{sec:synthetic_data}, the policy interacts with the environment and sequentially predicts decision-level action chunks,
\begin{equation}
a_t
\sim
\pi_\theta
\big(
\cdot
\mid
\mathcal{I},s_t
\big),
\qquad
\tau
=
\big(
(s_t,a_t)
\big)_{t=1}^{T}.
\label{eq:rl_trajectory}
\end{equation}
Here, $T$ is the actual number of model-level decisions before the episode terminates, distinct from the primitive-action budget $S_{\max}$. We use a binary task-level reward to make task completion the explicit objective, without requiring agreement with a reference action sequence. Success follows the stopping and distance criteria in Section~\ref{sec:overview}:$
R(\tau)
=
\mathbb{I}(\mathrm{success})
\in
\{0,1\}.
\label{eq:rl_reward}
$
We maximize the expected task reward over the trajectory distribution $p_\theta(\tau)$ induced by policy--environment interaction on synthesized episodes:$
\max_{\theta}\;
\mathcal{J}_{\mathrm{RL}}(\theta)
=
\mathbb{E}_{\tau\sim p_\theta}
\left[
R(\tau)
\right].
\label{eq:rl_objective}
$
Unlike imitation learning, this objective does not require agreement with a particular reference action sequence: any trajectory that successfully completes the task is rewarded. Training therefore progresses from expert action supervision, through policy-dependent supervised adaptation, to episode-level task-completion feedback.
\section{Experiments}
\label{sec:experiments}

\begin{table*}[t]
  \caption{\textbf{Comparison with state-of-the-art methods on inference
  latency and navigation performance.} Latency is measured on R2R-CE using the same NVIDIA A100 GPU, and measures model execution only for one action-chunk decision, excluding preprocessing, environment execution, and other non-model overhead. SimpleVLN uses Qwen3-VL-4B as its backbone.}
  \vspace{-1mm}
  \label{tab:inference_efficiency}
  \centering
  \footnotesize
  \setlength{\tabcolsep}{2.7pt}
  \begin{tabularx}{\textwidth}{lcc|*{4}{>{\centering\arraybackslash}X}|*{4}{>{\centering\arraybackslash}X}}
  \toprule
  \multirow[c]{2}{*}[-3pt]{Method}
  & \multirow[c]{2}{*}[-3pt]{Size}
  & \multirow[c]{2}{*}[-3pt]{\begin{tabular}[c]{@{}c@{}}Latency\\(ms)$\downarrow$\end{tabular}}
  & \multicolumn{4}{c|}{R2R-CE}
  & \multicolumn{4}{c}{RxR-CE} \\
  \cmidrule(lr){4-7}
  \cmidrule(lr){8-11}
  & &
  & NE$\downarrow$ & OS$\uparrow$ & SR$\uparrow$ & SPL$\uparrow$
  & NE$\downarrow$ & SR$\uparrow$ & SPL$\uparrow$ & nDTW$\uparrow$ \\
  \midrule

  StreamVLN \citep{streamvln} & 7B
  & 300
  & 4.98 & 64.2 & 56.9 & 51.9
  & 6.22 & 52.9 & 46.0 & 61.9 \\

  JanusVLN \citep{janusvln} & 7B
  & 906
  & 4.78 & 65.2 & 60.5 & 56.8
  & 6.06 & 56.2 & 47.5 & 62.1 \\

  DualVLN \citep{dualvln} & 7B
  & 683
  & 4.05 & 70.7 & 64.3 & 58.5
  & 4.58 & 61.4 & 51.8 & 70.0 \\

  FutureNav \citep{futurenav} & 4B
  & 1141
  & 4.24 & 70.2 & 65.4 & 61.3
  & 4.60 & 60.9 & 52.8 & 68.0 \\

  \midrule

  \rowcolor{gray!15}
  \textbf{SimpleVLN} & \textbf{4B}
  & \textbf{293}
  & \textbf{3.39} & \textbf{79.4} & \textbf{73.8} & \textbf{66.7}
  & \textbf{3.29} & \textbf{75.1} & \textbf{65.3} & \textbf{71.5} \\

  \bottomrule
  \end{tabularx}
  \vspace{-3mm}
\end{table*}
\begin{table*}[t]
\caption{\textbf{Navigation interface and memory design.} Ablations of action chunking, single-turn navigation, and observation sampling using Qwen3-VL 4B fine-tuned on R2R. The default design of SimpleVLN is shaded.}
\label{tab:formulation}
\vspace{-1mm}
\centering
\small

\begin{subtable}[t]{0.32\textwidth}
\vspace{0pt}
\centering
\footnotesize
\setlength{\tabcolsep}{2.8pt}
\begin{tabular}{lccc}
\toprule
Action format & NE$\downarrow$ & OS$\uparrow$ & SR$\uparrow$ \\
\midrule
Primitive action & 5.2 & 54.5 & 45.6 \\
\rowcolor{gray!15}
\textbf{Action chunk} & \textbf{5.5} & \textbf{61.2} & \textbf{53.6} \\
\bottomrule
\end{tabular}\par
\vspace{\baselineskip}
\caption{\textbf{Action chunking.} \emph{Primitive action} predicts one atomic action per decision; \emph{action chunk} predicts multiple atomic actions per decision and executes them in order, stopping early if the episode terminates.}
\label{tab:form_merge}
\end{subtable}
\hfill
\begin{subtable}[t]{0.32\textwidth}
\vspace{0pt}
\centering
\footnotesize
\setlength{\tabcolsep}{2.6pt}
\begin{tabular}{lccc}
\toprule
Dialogue format & NE$\downarrow$ & OS$\uparrow$ & SR$\uparrow$ \\
\midrule
Multi-turn & 5.9 & 55.2 & 35.7 \\
\rowcolor{gray!15}
\textbf{Single-turn} & \textbf{5.5} & \textbf{61.2} & \textbf{53.6} \\
\bottomrule
\end{tabular}\par
\vspace{\baselineskip}
\caption{\textbf{Single-turn navigation.} \emph{Multi-turn} accumulates observations and actions in the dialogue; \emph{single-turn} reconstructs each input from the instruction, and sampled fixed-budget global--local history.}
\label{tab:form_turn}
\end{subtable}
\hfill
\begin{subtable}[t]{0.32\textwidth}
\vspace{0pt}
\centering
\footnotesize
\setlength{\tabcolsep}{2.2pt}
\begin{tabular}{lccc}
\toprule
Sampling scheme & NE$\downarrow$ & OS$\uparrow$ & SR$\uparrow$ \\
\midrule
Chunk & 6.1 & 55.2 & 46.3 \\
Primitive (all) & 5.8 & 57.5 & 51.9 \\
\rowcolor{gray!15}
\textbf{Primitive (8)} & \textbf{5.5} & \textbf{61.2} & \textbf{53.6} \\
\bottomrule
\end{tabular}
\caption{\textbf{Observation sampling.} \emph{Chunk} samples boundaries; \emph{Primitive (all)} keeps all primitive-step observations; \emph{Primitive (8)} uniformly samples eight global history frames, excluding local frames.}
\label{tab:form_unit}
\end{subtable}
\vspace{-3mm}
\end{table*}

\subsection{Experimental Setup}
\label{sec:setup}

\noindent\textbf{Benchmarks and metrics.}
We evaluate SimpleVLN on the R2R-CE~\citep{r2r} and RxR-CE~\citep{rxr} val-unseen splits in Matterport3D with Habitat~\citep{vlnce}, reporting NE, OS, SR, SPL, and nDTW, with SR as the primary metric.

\noindent\textbf{Implementation details.}
We use Qwen3-VL (4B/8B)~\citep{qwen3vl} and Qwen3.5 (4B/9B)~\citep{qwen35}.
IL uses SFT in two stages. The first stage trains on the R2R~\citep{r2r} and RxR~\citep{rxr} demonstrations; the second stage initializes from the first-stage checkpoint and trains on the union of R2R, RxR, ScaleVLN [15], and the collected DAgger trajectories. The vision encoder is frozen in both stages.
Main-result RL uses DAPO~\citep{dapo} for 40 steps with batch size 256, 16 rollouts per prompt, learning rate $1\times10^{-6}$, KL coefficient $0$, and clip ratios $0.30/0.20$.

\subsection{Comparison with State-of-the-Art}
\label{sec:main_results}

\noindent\textbf{Main results.}
Table~\ref{tab:main} compares SimpleVLN with prior methods and reports results across multiple backbone families and model sizes, including pre-RL Qwen2.5-VL (3B/7B) variants. SimpleVLN reaches $76.7$ SR on R2R-CE and $80.1$ SR on RxR-CE using only single-RGB observations and the native MLLM architecture, outperforming existing methods. RL consistently improves SR across evaluated backbones; for Qwen3-VL 4B, SR increases from $65.4$ to $73.8$ on R2R-CE and from $63.4$ to $75.1$ on RxR-CE. Table~\ref{tab:inference_efficiency} shows that SimpleVLN requires $293$ ms per action-chunk decision, comparable to StreamVLN and faster than the other compared methods.
\subsection{Formulation}
\label{sec:interface_analysis}

We investigate the navigation formulation introduced in Section~\ref{sec:formulation}. Unless otherwise specified, these experiments use Qwen3-VL 4B fine-tuned on R2R and evaluated on the R2R-CE \textit{val-unseen} split.

\begin{figure}[t]
\centering
\includegraphics[width=0.48\textwidth]{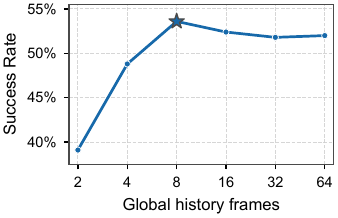}\hfill
\includegraphics[width=0.48\textwidth]{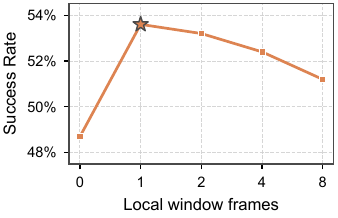}
\vspace{-1mm}
\caption{\textbf{Memory budget.} Effect of global and local memory sizes on R2R-CE \textit{val-unseen} SR. We vary $M$ with $N=1$ (left) and $N$ with $M=8$ (right); stars indicate the default configuration.}
\label{fig:abl_mem}
\vspace{-4mm}
\end{figure}

\noindent\textbf{Action chunking.} Table~\ref{tab:form_merge} evaluates action chunking as an alternative to primitive-action prediction. Action chunking raises SR from $45.6$ to $53.6$, an $8.0$-point gain, supporting it as an effective output representation beyond reducing model invocation frequency.

\noindent\textbf{Single-turn decision input.} Table~\ref{tab:form_turn} replaces accumulated multi-turn dialogue with a reconstructed single-turn input and sampled visual history, raising SR from $35.7$ to $53.6$. Although multi-turn dialogue can retain a fuller interaction history, continually appending observations and actions creates an overly long context that current MLLMs cannot effectively exploit for navigation.

\noindent\textbf{Observation sampling.} Table~\ref{tab:form_unit} examines which observations to retain. Sampling only chunk boundaries yields $46.3$ SR, retaining all primitive-step observations yields $51.9$, and uniformly sampling eight global history frames yields $53.6$. The eight-frame budget covers global history only, excluding local window. These results favor compact, uniformly sampled global history over retaining every frame.

\noindent\textbf{Memory budget.} Figure~\ref{fig:abl_mem} tests whether increasing the global or local history budget improves navigation accuracy. With $N=1$, SR peaks at $M=8$ and then declines as the global budget grows, suggesting that excessive global history introduces redundant or less relevant observations that dilute useful context. With $M=8$, retaining one local observation gives the highest SR, indicating that additional nearby frames provide largely redundant information. We therefore use $M=8$ and $N=1$: up to eight global and one local history frame.

\subsection{Data Construction}
\label{sec:data_analysis}

We investigate the effects of the training data constructed in Section~\ref{sec:synthetic_data}. All experiments in this subsection use Qwen3-VL 4B and report results on the R2R-CE and RxR-CE \textit{val-unseen} splits.

\begin{table*}[t]
\caption{\textbf{Demonstration and DAgger data ablation} on the R2R-CE and RxR-CE \textit{val-unseen} splits.}
\vspace{-1mm}
\label{tab:sft}
\centering
\scriptsize
\setlength{\tabcolsep}{5pt}
\resizebox{1.0\textwidth}{!}{%
\begin{tabular}{cccc cccc cccc}
\toprule
\multirow{2}{*}{R2R} & \multirow{2}{*}{RxR} & \multirow{2}{*}{ScaleVLN} &
\multirow{2}{*}{DAgger} & \multicolumn{4}{c}{R2R-CE} & \multicolumn{4}{c}{RxR-CE} \\
\cmidrule(lr){5-8}\cmidrule(lr){9-12}
 & & & & NE$\downarrow$ & OS$\uparrow$ & SR$\uparrow$ & SPL$\uparrow$
 & NE$\downarrow$ & SR$\uparrow$ & SPL$\uparrow$ & nDTW$\uparrow$ \\
\midrule
\checkmark &            &            &            & 5.53 & 61.2 & 53.6 & 48.1
& 6.45 & 50.3 & 42.0 & 57.6 \\
\checkmark & \checkmark &            &            & 5.28 & 63.4 & 56.7 & 51.3
& 5.75 & 54.5 & 46.7 & 62.4 \\
\checkmark & \checkmark & \checkmark &            & 4.90 & 68.4 & 57.9 & 50.9
& 5.27 & 55.3 & 48.1 & 63.6 \\
\rowcolor{gray!15}
\textbf{\checkmark} & \textbf{\checkmark} & \textbf{\checkmark} & \textbf{\checkmark} & \textbf{4.26} & \textbf{74.3} & \textbf{65.4} & \textbf{58.8}
& \textbf{4.79} & \textbf{63.4} & \textbf{54.1} & \textbf{66.5} \\
\bottomrule
\end{tabular}}
\vspace{-2mm}
\end{table*}


\noindent\textbf{Expert demonstration coverage.} Table~\ref{tab:sft} tests whether broadening expert demonstration coverage improves navigation. Adding RxR to R2R raises SR from $53.6$ to $56.7$ on R2R-CE and from $50.3$ to $54.5$ on RxR-CE. Including ScaleVLN further increases SR to $57.9$ and $55.3$, respectively. These results show that combining R2R and RxR provides clear in-domain gains, while ScaleVLN further improves generalization to unseen environments.

\noindent\textbf{Policy-dependent experience via DAgger.} We next assess whether policy-dependent data improves over the combined expert demonstrations. Adding DAgger yields the largest gains in Table~\ref{tab:sft}: $7.5$ SR points on R2R-CE and $8.1$ on RxR-CE, reaching $65.4$ and $63.4$. This shows the value of complementing static demonstrations with trajectories collected by the learned policy.

\noindent\textbf{Task resynthesis for RL.} Table~\ref{tab:rl} evaluates RL tasks synthesized from the R2R, RxR, and ScaleVLN training scenes. These names denote scene pools for synthesis, not reused annotated episodes. All variants start from the same IL--DAgger checkpoint and use GRPO~\citep{grpo} with batch size 32 and eight rollouts per prompt. Each pool improves SR over the pre-RL baseline, while their mixture performs best, reaching $70.1$ on R2R-CE and $71.0$ on RxR-CE.

\begin{table*}[t]
\caption{\textbf{Synthesized RL data ablation} on the R2R-CE and RxR-CE \textit{val-unseen} splits. Dataset columns identify the training-scene sources used to synthesize RL episodes.}
\vspace{-1mm}
\label{tab:rl}
\centering
\scriptsize
\setlength{\tabcolsep}{7pt}
\resizebox{1.0\textwidth}{!}{%
\begin{tabular}{ccc cccc cccc}
\toprule
\multirow{2}{*}{R2R} & \multirow{2}{*}{RxR} & \multirow{2}{*}{ScaleVLN}
& \multicolumn{4}{c}{R2R-CE} & \multicolumn{4}{c}{RxR-CE} \\
\cmidrule(lr){4-7}\cmidrule(lr){8-11}
 & & & NE$\downarrow$ & OS$\uparrow$ & SR$\uparrow$ & SPL$\uparrow$
 & NE$\downarrow$ & SR$\uparrow$ & SPL$\uparrow$ & nDTW$\uparrow$ \\
\midrule
\multicolumn{3}{c}{Pre-RL baseline} & 4.26 & 74.3 & 65.4 & 58.8
& 4.79 & 63.4 & 54.1 & 66.5 \\
\midrule
\checkmark &            &            & 3.91 & 74.0 & 68.5 & 63.2
& 4.31 & 65.8 & 55.3 & 68.3 \\
 & \checkmark &            & 3.76 & 71.8 & 67.5 & 59.3
& 3.98 & 68.5 & 59.5 & 68.8 \\
           &            & \checkmark & 4.17 & 70.9 & 66.6 & 61.3
& 4.34 & 66.3 & 56.2 & 67.3 \\
\checkmark & \checkmark &            & 3.75 & 74.1 & 69.4 & 63.8
& 3.77 & 69.7 & 60.8 & 68.1 \\
\rowcolor{gray!15}
\textbf{\checkmark} & \textbf{\checkmark} & \textbf{\checkmark} & \textbf{3.74} & \textbf{73.1} & \textbf{70.1} & \textbf{65.6} & \textbf{3.82} & \textbf{71.0} & \textbf{62.1} & \textbf{68.4} \\
\bottomrule
\end{tabular}}
\vspace{-3mm}
\end{table*}

\begin{table*}[t]
\caption{\textbf{Reward design and RL optimization} on the R2R-CE and RxR-CE \textit{val-unseen} splits with Qwen3-VL 4B. The SPL reward is zero for failed episodes and equals the ratio of the ground-truth path length to the executed path length for successful episodes; the success reward is binary.}
\vspace{-1mm}
\label{tab:reward}
\centering
\scriptsize
\setlength{\tabcolsep}{9pt}
\resizebox{\textwidth}{!}{%
\begin{tabular}{l cccc cccc}
\toprule
\multirow{2}{*}{RL configuration} & \multicolumn{4}{c}{R2R-CE}
& \multicolumn{4}{c}{RxR-CE} \\
\cmidrule(lr){2-5}\cmidrule(lr){6-9}
& NE$\downarrow$ & OS$\uparrow$ & SR$\uparrow$ & SPL$\uparrow$
& NE$\downarrow$ & SR$\uparrow$ & SPL$\uparrow$ & nDTW$\uparrow$ \\
\midrule
Pre-RL baseline & 4.26 & 74.3 & 65.4 & 58.8 & 4.79 & 63.4 & 54.1 & 66.5 \\
\midrule
GRPO + $\mathrm{SPL}(\tau)$ & 3.80 & 73.9 & 68.9 & 63.1
& 4.13 & 67.7 & 58.3 & 67.3 \\
GRPO + $\mathbb{I}(\mathrm{success})$
& 3.74 & 73.1 &
70.1 & 65.6
& 3.82 & 71.0 & 62.1 & 68.4 \\
\rowcolor{gray!15}
\textbf{DAPO + $\mathbb{I}(\mathrm{success})$} & \textbf{3.45} & \textbf{76.6} & \textbf{71.5} & \textbf{65.3}
& \textbf{3.39} & \textbf{73.7} & \textbf{64.0} & \textbf{69.4} \\
\bottomrule
\end{tabular}}
\vspace{-5mm}
\end{table*}

\begin{figure}[!h]
\centering
\begin{subfigure}[t]{0.315\textwidth}
  \centering
  \includegraphics[width=\linewidth]{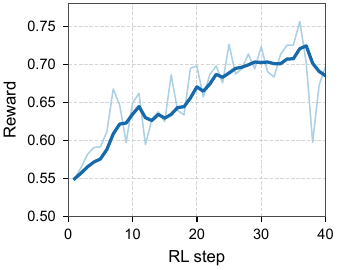}
  \vspace{-6mm}
  \caption{Rollout reward.}
  \label{fig:rl_reward}
\end{subfigure}
\hfill
\begin{subfigure}[t]{0.315\textwidth}
  \centering
  \includegraphics[width=\linewidth]{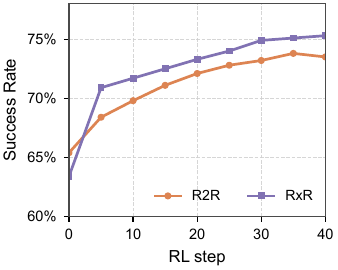}
  \vspace{-6mm}
  \caption{Success rate on R2R and RxR.}
  \label{fig:rl_sr}
\end{subfigure}
\hfill
\begin{subfigure}[t]{0.315\textwidth}
  \centering
  \includegraphics[width=\linewidth]{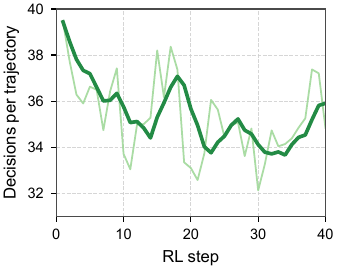}
  \vspace{-6mm}
  \caption{Decisions per trajectory.}
  \label{fig:rl_length}
\end{subfigure}
\vspace{-2mm}
\caption{\textbf{RL training dynamics.} Overall, rollout reward and evaluation success rate increase, while the average number of action-chunk decisions per trajectory decreases.}
\label{fig:rl_dynamics}
\vspace{-5mm}
\end{figure}

\subsection{Optimization}
\label{sec:optimization_analysis}

We investigate how RL configurations affect navigation accuracy and how the model's navigation behavior evolves during training. Table~\ref{tab:reward} uses Qwen3-VL 4B and a mixture of synthetic episodes constructed from the R2R, RxR, and ScaleVLN training scenes.

\noindent\textbf{Reward design.}
Under GRPO with batch size 32 and eight rollouts per prompt, we compare SPL and binary rewards. Both assign zero to failures; successful episodes receive either the SPL score or a reward of one. The binary reward raises SR from $68.9$ to $70.1$ on R2R-CE and from $67.7$ to $71.0$ on RxR-CE, supporting simple task-completion feedback.

\noindent\textbf{RL configuration.}
With the success reward fixed, DAPO~\citep{dapo} reaches $71.5/73.7$ SR, compared with $70.1/71.0$ for GRPO. Both ablations use a batch size of 32, compared with 256 for the main-results configuration in Section~\ref{sec:setup}. These results favor the evaluated DAPO configuration, although we do not attribute the gain solely to the optimization algorithm.

\noindent\textbf{RL training dynamics.}
Figure~\ref{fig:rl_dynamics} shows that, despite fluctuations, rollout reward and evaluation SR trend upward, while the average number of action-chunk decisions per trajectory gradually decreases. This suggests that RL improves navigation performance while also increasing decision efficiency.

\begin{figure}[!h]
\centering
\includegraphics[width=0.98\textwidth]{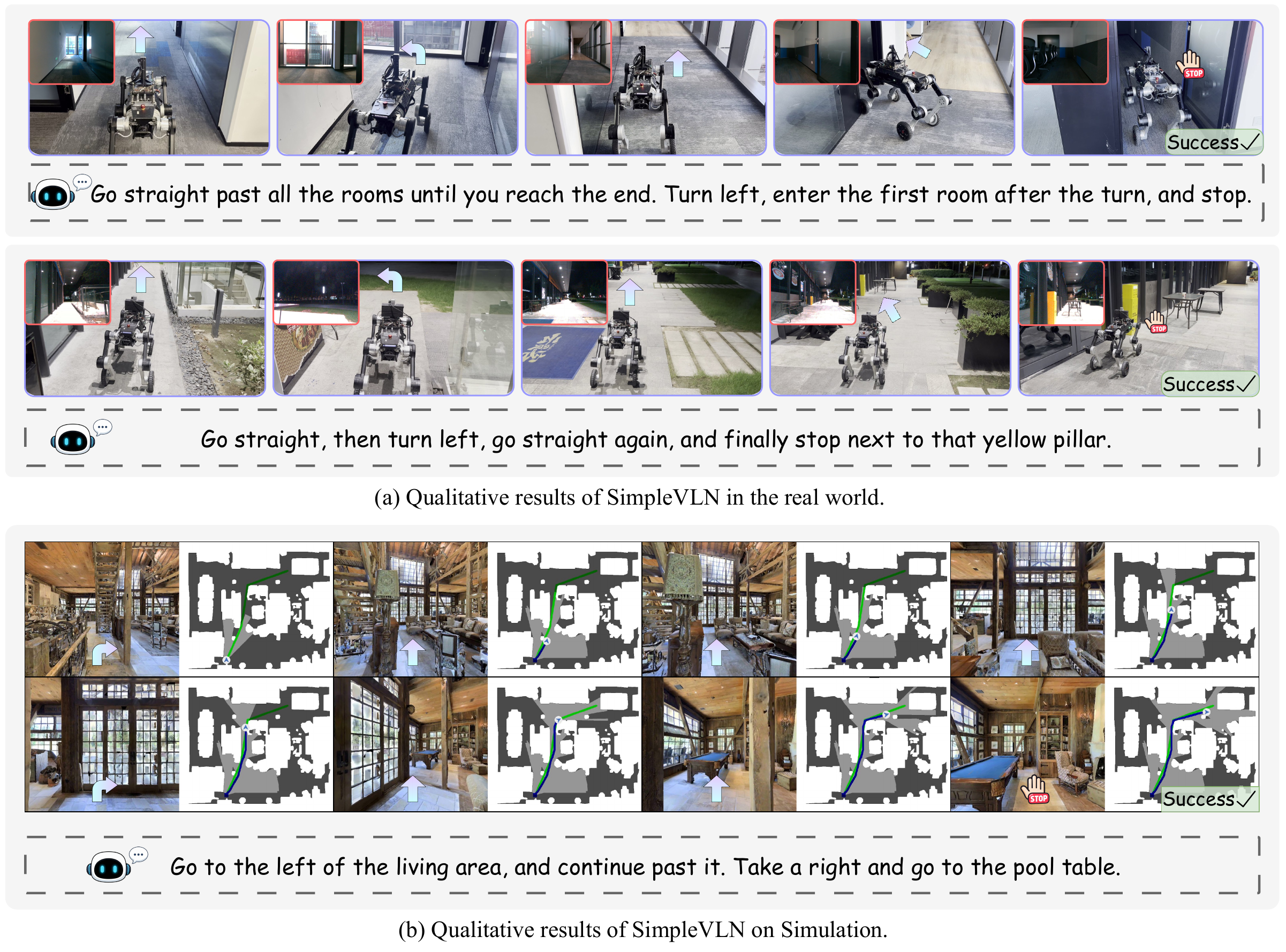}
\vspace{-2mm}
\caption{\textbf{SimpleVLN in real-world and simulated environments.}
SimpleVLN follows diverse instructions and reaches the specified goals; arrows indicate navigation directions.}
\label{fig:sim_and_real}
\vspace{-5mm}
\end{figure}

\subsection{Real-world Experiments}
We further evaluate SimpleVLN in real environments   using a Daystar quadruped robot.
The model runs onboard on an NVIDIA Jetson Thor GPU. The robot is equipped with an Intel
RealSense D450 RGB-D camera for egocentric perception. We only use RGB as our observations. Figure \ref{fig:sim_and_real}  shows that SimpleVLN successfully follows multi-step instructions involving forward movement, turns, room entry, and landmark-based stopping, while maintaining coherent instruction-following behavior over extended trajectories.

\vspace{-2mm}
\section{Conclusion}
\label{sec:conclusion}
\vspace{-2mm}

We presented SimpleVLN, a framework for adapting native MLLMs to VLN. SimpleVLN combines single-turn decision inputs, global--local memory, and action chunking with training data constructed from expert demonstrations, policy-dependent DAgger experience, and synthesized RL tasks. A progressive IL--DAgger--RL pipeline further shifts learning from action imitation to task-level outcome optimization. SimpleVLN achieves strong navigation performance, demonstrating that native MLLM navigation capabilities can be effectively unlocked without task-specific architectural designs.

{
\small
\bibliographystyle{plain}
\bibliography{example}
}


\appendix


\end{document}